\documentclass[final]{cvpr}

\usepackage{times}
\usepackage{epsfig}
\usepackage{graphicx}
\usepackage{amsmath}
\usepackage{amssymb}

\usepackage{bm}
\usepackage{booktabs}
\usepackage{multirow}
\usepackage[table]{xcolor}
\definecolor{yelloworange}{RGB}{255, 153, 0}

\def\0{{\mathbf 0}}
\def\1{{\mathbf 1}}

\newcommand{\tabincell}[2]{\begin{tabular}{@{}#1@{}}#2\end{tabular}}
\def\Name{LiDAR R-CNN }
\def\NameNS{LiDAR R-CNN}

\graphicspath{{./images/}}

\usepackage[pagebackref=false,breaklinks=true,letterpaper=true,colorlinks,bookmarks=false]{hyperref}
\def\cvprPaperID{1579} 
\def\confYear{CVPR 2021}

\begin{document}

\title{LiDAR R-CNN: An Efficient and Universal 3D Object Detector}

\author{
	Zhichao Li*\quad
	Feng Wang*\quad
	Naiyan Wang\\
	TuSimple\\
	{\tt\small
		\{leeisabug, feng.wff, winsty\}@gmail.com
	}
}

\maketitle
{\let\thefootnote\relax\footnote{*The first two authors contribute equally to this work.}}

\begin{abstract}
	LiDAR-based 3D detection in point cloud is essential in the perception system of autonomous driving. 
	In this paper, we present {\NameNS}, a second stage detector that can generally improve any existing 3D detector. To fulfill the real-time and high precision requirement in practice, we resort to point-based approach other than the popular voxel-based approach. However, we find an overlooked issue in previous work: Naively applying point-based methods like PointNet could make the learned features ignore the size of proposals.
	To this end, we analyze this problem in detail and propose several methods to remedy it, which bring significant performance improvement. Comprehensive experimental results on real-world datasets like Waymo Open Dataset (WOD) and KITTI dataset with various popular detectors demonstrate the universality and superiority of our \NameNS. In particular, based on one variant of PointPillars, our method could achieve new state-of-the-art results with minor cost. Codes will be released at \url{https://github.com/tusimple/LiDAR_RCNN}.
	
\end{abstract}
\thispagestyle{empty}
\thispagestyle{empty}
\section{Introduction}
For autonomous vehicles and robots, estimating the 7 Degrees-of-Freedom (location, dimension, and orientation) state of the surrounding objects in complicated real-world environments is a vital task. Recently, LiDAR-based 3D object detection has been received increasing attention~\cite{pv-rcnn,votenet,clocs} due to its ability of direct 3D measurement. However, compared with the well developed 2D image detection, LiDAR-based 3D detection still suffers from the intrinsic difficulties of point sparsity and large search space in 3D space. 

Given that point clouds from LiDAR are irregular, most 3D detection methods transform such data into regular 3D voxel grids~\cite{song2016deep, voxelnet,second,pointpillars} or collections of projected 2D view~\cite{joint3d, mv3d, pixor, conticonv, hdnet, mmf, rangercnn}. While these methods can easily take advantage of the ordered data representation by using regular 2D or 3D convolution for feature extraction conveniently, the quantization error in the construction of voxel or multi-view features limits their performance. On the contrary, point-based methods~\cite{pointnet++,pointrcnn,frustumconv, frustum, std} can learn features from the raw point cloud, but usually, they need complicated and inefficient operations~\cite{pointnet++} to aggregate local information. Consequently, these point-based methods are mostly fused with other representations other than individually used for 3D object detection. 

In this paper, we are more interested in another setting like R-CNN~\cite{rcnn}: We already have a set of proposals in 3D space and seek to refine them. Therefore, we name our method {\NameNS}. When confronting this usage, PointNet becomes our first choice since the network only needs to process the points from a single object in a small region. Compared with the DNN features with rich semantic information, the original point cloud contains the most accurate location information. 
Intuitively, applying PointNet~\cite{pointnet} on the raw point cloud for detection is straightforward. However, we find that the results of such a naive implementation are unsatisfactory. Through careful analysis,
we find an intriguing size ambiguity problem: Different from voxel-based methods that equally divide the space into fixed size, PointNet only aggregates the features from points while ignoring the spacing in 3D space. Nevertheless, the spacing encodes essential information, such as the scale of the object.
To remedy this issue, we propose a series of solutions and prove their effectiveness through detailed experiments.
Comprehensive evaluation results on the Waymo Open Dataset(WOD)~\cite{wod}
prove that our proposed \Name can improve the performance of various off-the-shelf detectors significantly and consistently. With a well-tuned PointPillars model, we even outperform the state-of-the-art models. 


To summarize, our contributions are as follows:
\begin{itemize}
	\item We propose an R-CNN style second-stage detector for 3D object detection based on PointNet. Our method is plug-and-play to any existing 3D detector, and needs no re-training to the base detectors.
	\item We reveal the size ambiguity problem when using a point-based R-CNN detector. Through careful analysis, we propose several different ways to make the detector aware of the size of proposal box. Despite the simple design, it achieves significant performance improvements.
	\item We test our proposed method on WOD~\cite{wod} and KITTI~\cite{kitti} datasets with various base detectors. Our method could consistently improve the base detectors while running at 200fps for 128 proposals on 2080Ti GPU.  
\end{itemize}

\section{Related Work}

This section will briefly review some closely related work in LiDAR 3D detection according to different data representations.
\subsection{Voxel based Methods}
Researchers have long struggled with the irregular organization of point clouds. This nature makes it infeasible to apply traditional convolution, which is defined on regular grid. Voxelization~\cite{voxelfpn, class} is one of the most common treatments for it. ~\cite{wang2015voting, vote3deep} perform real-time 3D detection by converting point cloud data into voxels with hand-crafted features. Nevertheless, the generalization of hand-crafted features limits their performance in complicated real-world environments. Subsequent work improves them in various ways: ~\cite{3dconv} introduces a 3D fully convolutional network which uses binary 3D voxel representation. VoxelNet~\cite{voxelnet} applies mini PointNet for discriminative voxel features extraction. SECOND~\cite{second} further overcomes the computational barrier of dense 3D convolution by applying sparse convolution. ~\cite{hansong3d} proposes 3D neural architecture search (3D-NAS) to search for efficient 3D models structures.
Albeit these voxel-based methods are widely used, their performance upper bounds are still limited by the quantization error when dividing the voxels. Thus voxel-based methods are usually fused with other representations that will be introduced later.

\subsection{View based methods}
CNN-based networks, such as VGGNet~\cite{vgg}, ResNet~\cite{resnet}, etc., have demonstrated excellent feature extraction ability on 2D computer vision tasks. An intuitive idea is to transform the point cloud data into certain kind of 2D view, which could be efficiently processed by 2D CNN. Among them, bird-eye-view (BEV)~\cite{yolo3d, joint3d, afdet, polarnet, ssn} is one of the most common representations. PointPillars~\cite{pointpillars} collapses the points in vertical columns (pillars). MV3D~\cite{mv3d} combines features from BEV and frontal view. While BEV could encode the scene to 2D space while maintaining scale invariance, range view~\cite{lasernet, dmlo,baidurange,squeezeseg, rangenet++, squeezesegv3, cylinder3d} is another compact representation based on the intrinsic 2.5D LiDAR scan. The range view avoids the challenge of the sparsity of point cloud yet introduces the scale variation for objects at different range. Recently, RCD~\cite{rcd,fan2021rangedet} tries to provide a uniform feature in 3D space with a range conditioned dilated convolution. And Soft Range Gating is used to avoid the influence of significant change in disparity. Benefited from the mature 2D CNN, view-based methods are usually faster, but as mentioned, they usually introduce information loss when projecting 3D space into 2D.

\subsection{Point based methods}
PointNet and its variants~\cite{pointnet, pointnet++, dgcnn, rscnn, pseudo} could directly take the raw points as input and use symmetric operators to address the main issue in points cloud -- unorderness. Based on these powerful tools, PointRCNN~\cite{pointrcnn} proposes a bottom-up 3D Region proposal network. They first generate proposals via segmented foreground objects and then learn a more accurate bounding box regression branch to predict the box coordinates. $\text{Part A}^2$~\cite{parta2} extends PointRCNN by introducing an intra-object part supervision.
PV-RCNN~\cite{pv-rcnn} aggregates points feature into a voxel-based framework, which improves the performance remarkably. From another perspective, VoteNet~\cite{votenet, imvotenet} proposes a new proposal generation mechanism with deep Hough voting based on the observation that the LiDAR points are all distributed on the surface of objects. StarNet~\cite{starnet} uses sampling centers with anchors as proposals to achieve a more computationally efficient framework, yet the low quality of the proposals limits its performance.

For a widely used Velodyne LiDAR HDL-64E, it collects more than 100K points in one scan. This is a considerable challenge for PointNet. So almost all the point-based methods need to downsample when dealing with large scenes. For PointRCNN, 16,384 points are kept. Besides, to effectively extract features from these individual points, usually hierarchical grouping methods as in ~\cite{pointnet++} are essential in these methods; however, they are usually costly compared with pure PointNet. To summarize, these two main issues limit the performance and efficiency of point based method in general 3D detection task.

\section{Methods}

In this section, we will firstly introduce our framework of point-based R-CNN model. Then we will describe the size ambiguity problem, which is the core issue that limits the performance. Next, we propose several simple yet effective solutions to solve it.

\subsection{Point-based R-CNN}

\begin{figure}[t]
	\centering
	\includegraphics[width=0.9\linewidth]{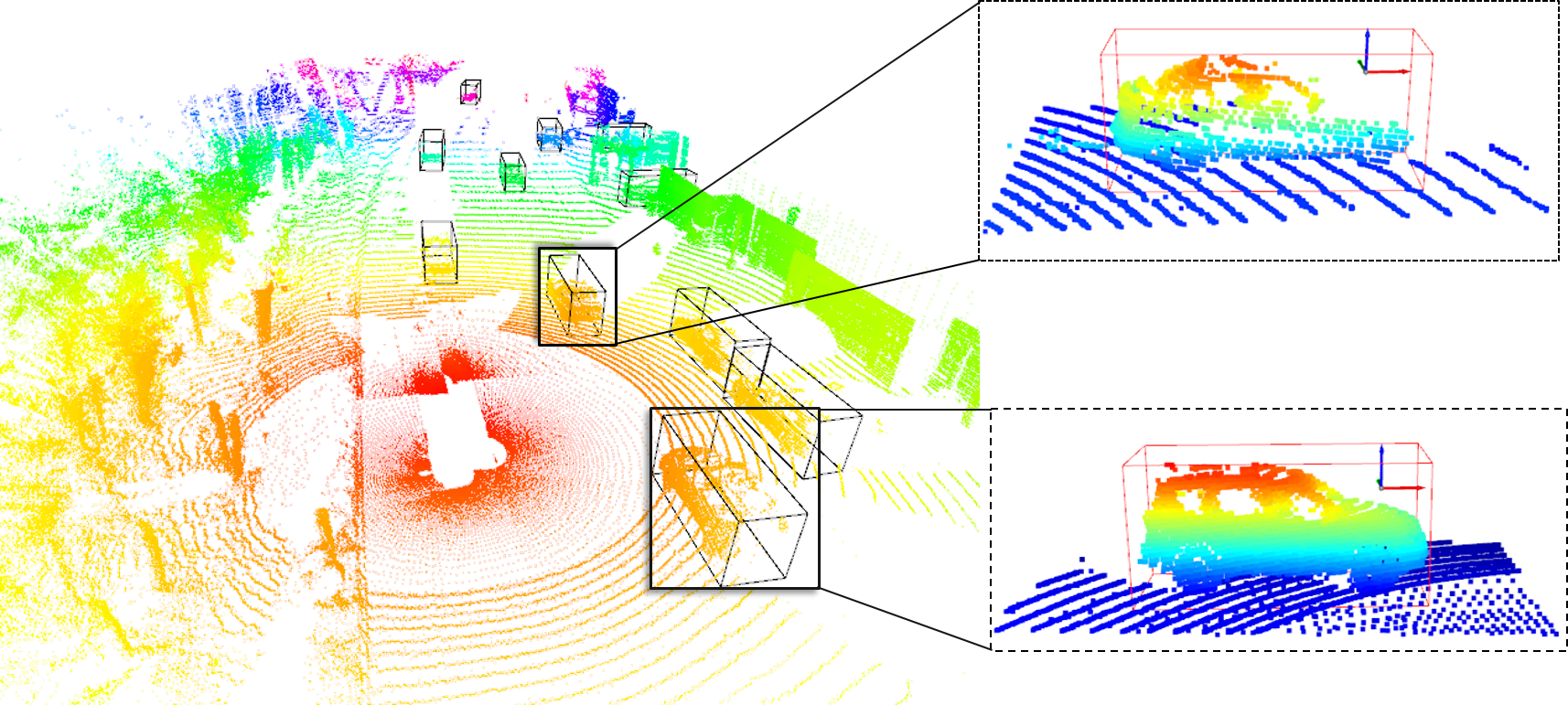}
	\vspace{1mm}
	\caption{Illustration of input features. We convert the points into the proposal’s coordinate system. 
		In this way, no matter where the object is, it is similar to the network, which improves the generalization ability. The proposal boxes are enlarged to contain some contextual points (usually ground points).}
	\label{fig:canonical}
	\vspace{-5mm}
\end{figure}

Assume that we already have a 3D LiDAR object detector (no matter deep learning based or not) that could generate lots of 3D bounding box proposals, our goal is to refine the 7DoF bounding box parameters (including size, position and heading) and scores of the proposals simultaneously, i.e. in an R-CNN manner. To make our model plug-and-play and avoid further quantization error or interpolation, our input data are the original point cloud without any high-level DNN features. 

\textbf{Input Features.}
For each 3D proposal $\mathbf{b}_i = (x_i, y_i, z_i, w_i, l_i, h_i, \theta_i)$, we enlarge its width and length to contain more contextual points around it. 
Then all the points in the enlarged boxes are taken out as the input data for our R-CNN model.

To improve the generalization of our R-CNN model, the points are then normalized according to the 3D bounding box coordinate system. The origin point is set as the center of the box. The heading direction is set as the $x$-axis. Its horizontally orthogonal direction is the $y$-axis and the vertical up direction is the $z$-axis (Fig. \ref{fig:canonical}).

\begin{figure}[t]
	\centering
	\includegraphics[width=0.9\linewidth]{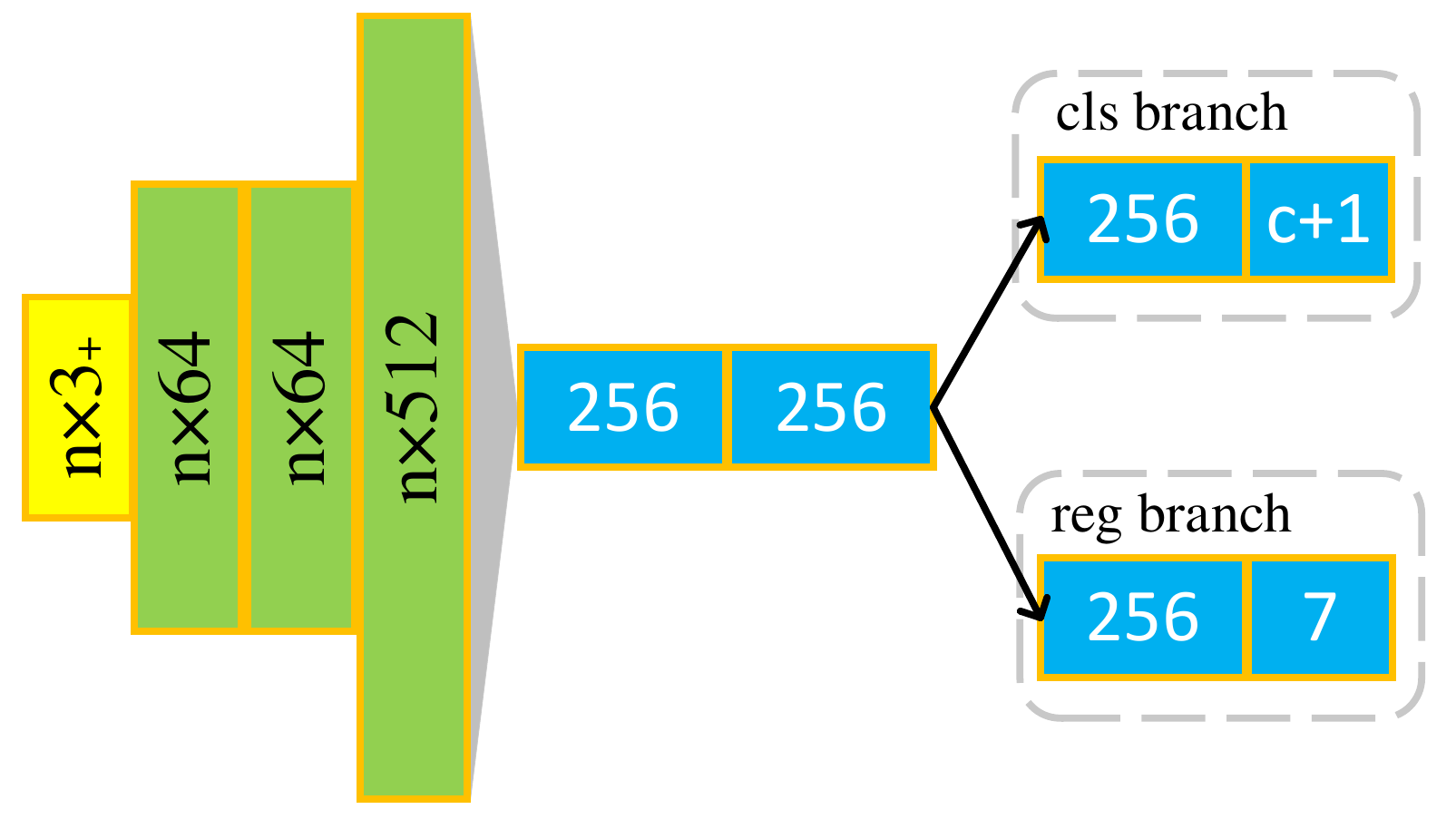}
	\caption{ The PointNet\cite{pointnet} structure for our R-CNN model. depending on different solutions, the input data has $3$ or more channels for each points. $c$ in the classification branch denotes the class number and $c+1$ includes an additional background class.}
	\label{fig:pointnet_backbone}
	\vspace{-3mm}
\end{figure}

\textbf{Backbone.}
The normalized points' coordinates $(x^p, y^p, z^p)$ and other meta-data will be mentioned later are fed into a point-based network. For fast inference, we choose PointNet\cite{pointnet} as our backbone network. It consists of a Multi-Layer Perception (MLP) module with three fully connected layers and a max-pooling operator for feature aggregation. Then the aggregated feature is fed into two branches: one for classification and another for regression. The whole network structure is illustrated in Fig \ref{fig:pointnet_backbone}. This backbone network is extremely light-weighted and fast, which is very suitable for real-time applications.

\textbf{Loss Function.}
Similar to 2D RCNN, the loss function for classification branch is a softmax cross entropy loss with $C+1$ categories,
\begin{equation}
	\mathcal{L}_{cls} = \frac{1}{B} \sum_{i=1}^{B} -\log p_{y(i)}^{(i)},
\end{equation}
where $C$ is the class number, $B$ is the batch size, $y(i)$ is the assigned label for $i$-th sample and $p_{y(i)}$ is the softmax probability on $y(i)$. The IoU threshold for positive and negative samples depends on the specific category, e.g. 0.7 for vehicle and 0.5 for pedestrian on WOD.

The regression loss aims at refining the box parameters, thus it is only applied to the positive samples. Since the boxes are already transformed to the bounding box coordinate system, the proposal $\mathbf{b}_i = (x_i, y_i, z_i, w_i, l_i, h_i, \theta_i)$ in global reference frame is transformed to
\begin{equation}
	\mathbf{\tilde{b}}_i = (0, 0, 0, w_i, l_i, h_i, 0).
\end{equation}
Meanwhile, the 3D ground-truth bounding box $\mathbf{b}_i^{gt} = (x_i^{gt}, y_i^{gt}, z_i^{gt}, w_i^{gt}, l_i^{gt}, h_i^{gt}, \theta_i^{gt})$ should also be transformed to
\begin{equation}
		\mathbf{\tilde{b}}_i^{gt} = (x_i^{gt} - x_i, y_i^{gt} - y_i, z_i^{gt} - z_i, w_i^{gt}, l_i^{gt}, h_i^{gt}, \theta_i^{gt} - \theta_i).
\end{equation}

\begin{figure*}[!t]
	\centering
	\includegraphics[width=0.9\linewidth]{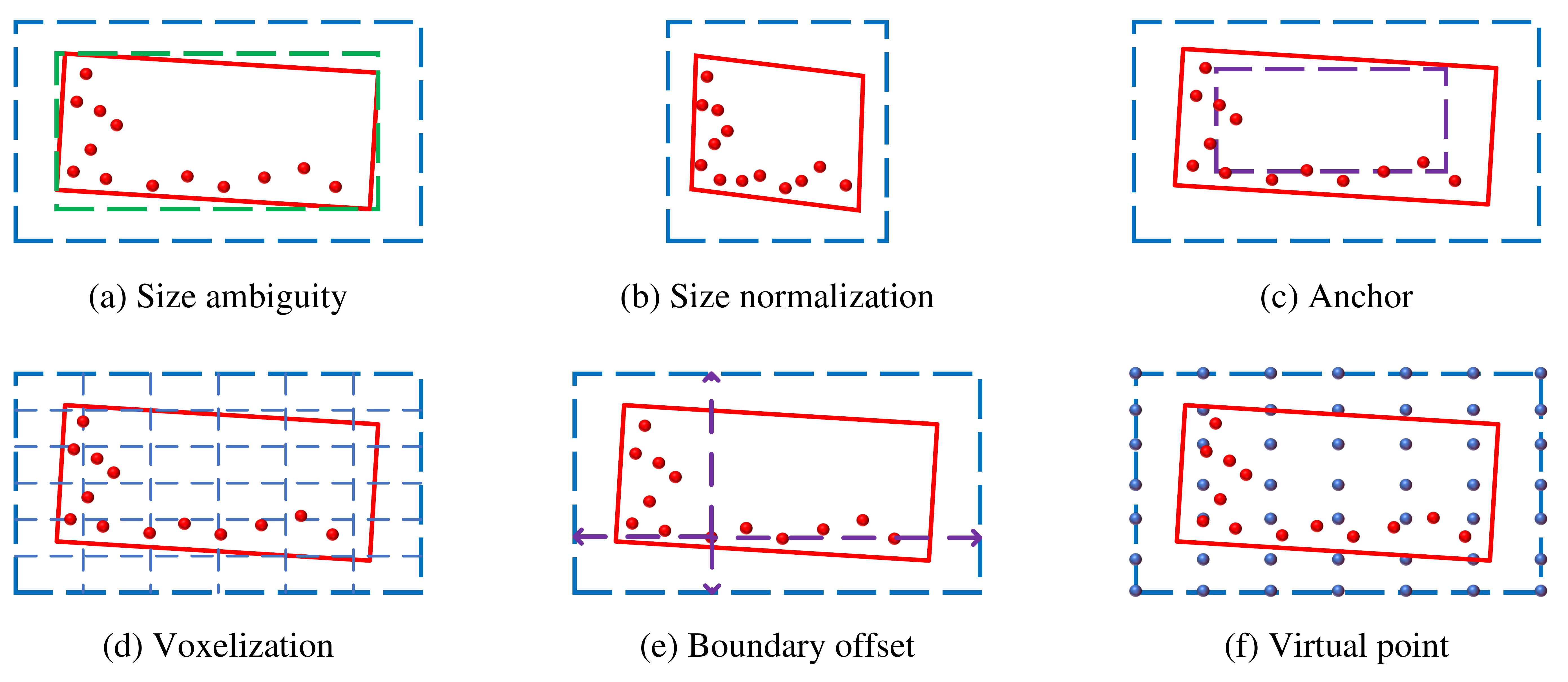}
	\caption{(a) Through point pooling operator, the two dashed bounding boxes extract the same points. However, the two dashed bounding boxes have different IoUs with the ground-truth (red box), which would confuse the R-CNN module. (b) Size normalization changes the shape of the point cloud. The rectangle ground-truth box becomes a parallelogram. (c) An anchor (purple box) for each category cannot provide the position information of the proposal. (d) Through voxelization, the model may know the box is enlarged because the border voxels have no points in them. However, it still doesn't know where the proposal's border is. (e) The boudnary offset information is directly concatenated with the point features. (f) By adding virtual grid points, the R-CNN model will be able to know the proposal is too large because the border virtual points have no real points in their neary neighbors. The virtual points also provide clear boundary information of the proposal. }
	\label{fig:solutions}
	\vspace{-5mm}
\end{figure*} 

Also similar to 2D RCNN, the regression targets for center and size are defined as
\begin{equation}
		\mathbf{t}_i^{c} = (\frac{x_i^{gt} - x_i}{w_i}, \frac{y_i^{gt} - y_i}{l_i}, \frac{z_i^{gt} - z_i}{h_i}),
\end{equation}
\begin{equation}
\label{eq:size_target}
	\mathbf{t}_i^{s} =(\log \frac{w_i^{gt}}{w_i}, \log \frac{l_i^{gt}}{l_i}, \log \frac{h_i^{gt}}{h_i}).
\end{equation}
For the orientation regression which is specific for 3D detection, we cannot directly make the target as $\theta_i^{gt} - \theta$. Due to the sparsity of point clouds and the appearance ambiguity of some objects, their predicted orientation may be flipped with $180^{\circ}$. If we simply use $\theta_i^{gt} - \theta$ as the target, the gradient of flipped samples will be too large to overwhelm the normal samples. 
To avoid the training contaminated by these outlier samples, we set the orientation target to be the minimum residual calculated by original orientation and flipped orientation:
\begin{equation}
	\Delta \theta_i = (\theta_i^{gt} - \theta_i) \ mod \  \pi,
\end{equation}
\begin{equation}
	\begin{aligned}
		\mathbf{t}_i^{o} = \left\{
			\begin{array}{ll}
			\Delta \theta_i , &\Delta \theta_i \leq \frac{\pi}{2}, \\
			\Delta \theta_i - \pi, & \Delta \theta_i > \frac{\pi}{2}.
		    \end{array} 
	        \right.
	\end{aligned}
\end{equation}

Having all the targets $\mathbf{t}_i = (\mathbf{t}_i^{c}, \mathbf{t}_i^{s}, \mathbf{t}_i^{o})$, our regression loss is defined as,
\begin{equation}
	\mathcal{L}_{reg} = \frac{1}{B_+} \sum_{i=1}^{B_+}SmoothL1(\mathbf{o}_i - \mathbf{t}_i),
\end{equation}
where $\mathbf{o}_i$ is the output of the model's regression branch and $B_+$ is the number of positive samples. Finally, the overall loss is formulated as
\begin{equation}
	\mathcal{L} = \mathcal{L}_{cls} + \lambda \mathcal{L}_{reg},
\end{equation}
Where $\lambda$ is a balance factor, which is 20 in practice.

\begin{table*}
	\small
	\begin{center}
		\scalebox{0.79}[0.79]{
			\setlength\tabcolsep{2pt}
			\begin{tabular}{c|c||cccc|cccc|cccc|cccc}
				\hline
				&  &
				\multicolumn{4}{c|}{3D AP (IoU=0.7)} & \multicolumn{4}{c|}{3D APH (IoU=0.7)} &
				\multicolumn{4}{c|}{BEV AP (IoU=0.7)} &
				\multicolumn{4}{c}{BEV APH (IoU=0.7)} \\
				\multirow{-2}{*}{Difficulty} & \multirow{-2}{*}{Method} & Overall & 0-30m & 30-50m & 50m-Inf &
				Overall & 0-30m & 30-50m & 50m-Inf  &
				Overall & 0-30m & 30-50m & 50m-Inf &
				Overall & 0-30m & 30-50m & 50m-Inf \\
				\hline
				LEVEL\_1
				& StarNet~\cite{starnet}  & 53.7 & - & - & - & - & - & - & - & - & - & - & - & - & - & - & -  \\
				\rowcolor{gray!20}
				& PointPillar~\cite{pointpillars}  & 56.6 & 81.0 & 51.8 & 27.9 & - & - & - & - & 75.6 & 92.1 & 74.1 & 55.5 & - & - & - & -  \\
				& MVF~\cite{mvf}  & 62.9 & 86.3 & 60.0 & 36.0 & - & - & - & - & 80.4 & 93.6 & 79.2 & 63.1 & - & - & - & -\\
				\rowcolor{gray!20}
				& Pillar-od~\cite{pillar-od}  & 69.8 & 88.5 & 66.5 & 42.9 & - & - & - & - & 87.1 & 95.8 & 84.7 & 72.1 & - & - & - & -\\
				& PV-RCNN~\cite{pv-rcnn} & 70.3 & 91.9 & 69.2 & 42.2 & 69.7 & 91.3 & 68.5 & 41.3 & 83.0 & \textbf{97.4} & 83.0 & 65.0 & 82.1 & \textbf{96.7} & 82.0 & 63.2 \\
				\cline{2-18}
				& RCD ~\cite{rcd} & 66.4 & 86.6 & 65.6 & 40.0 & 66.1 & 86.3 & 65.3 & 39.9 & 82.1 & 93.3 & 80.9 & 67.2 & 81.4 & 92.8 & 80.2 & 66.2 \\
				\rowcolor{gray!20}
				& RV first stage & 53.4 & 73.0 & 49.0 & 28.1 & 52.8 & 72.3 & 48.4 & 27.8 & 70.5 & 86.2 & 68.5 & 51.9 & 69.5 & 85.2 & 67.6 & 51.0 \\
				& \Name(rv) & 68.7 & 84.8 & 67.6 & 47.3 & 68.2 & 84.3 & 67.1 & 46.6 & 81.1 & 90.5 & 80.5 & 68.9 & 80.4 & 89.9 & 79.8 & 67.8 \\
				\cline{2-18}
				& PointPillars* & 72.1 & 88.3 & 69.9 & 48.0 & 71.5 & 87.8 & 69.3 & 47.3 & 87.9 & 96.6 & 87.1 & 78.1 & 87.1 & 96.0 & 86.2 & 76.5 \\
				\rowcolor{gray!20}
				& \Name(pp) & 75.6 &92.1 & 74.3 & 53.3 & 75.1 & 91.6 & 73.8 & 52.6 & 88.2 & 97.1 & 87.6 & 78.3 & 87.5 & 96.6 & 86.9 & 76.8 \\
				& \Name(2x) & 75.6 & 91.9 & 74.2 & 53.5 & 75.1 & 91.5 & 73.6 & 52.7 & 90.0 & 97.0 & 89.4 & 78.5 & 89.2 & 96.6 & 88.6 & 77.0 \\
				\rowcolor{gray!20}
				& \Name(2xc) & \textbf{76.0} &  \textbf{92.1} & \textbf{74.6} & \textbf{54.5} & \textbf{75.5} & \textbf{91.6} & \textbf{74.1} & \textbf{53.4} & \textbf{90.1} & 97.0 & \textbf{89.5} & \textbf{78.9} & \textbf{89.3} & 96.5 & \textbf{88.6} & \textbf{77.4} \\
				\hline
				LEVEL\_2 & PV-RCNN~\cite{pv-rcnn} & 65.4 & \textbf{91.6} & 65.1 & 36.5 & 64.8 & \textbf{91.0} & 64.5 & 35.7 & 77.5 & \textbf{94.6} & 80.4 & 55.4 &76.6 & \textbf{94.0} & 79.4 & 53.8\\
				\cline{2-18}
				& RV first stage & 46.0 & 70.6 & 44.1 & 21.0 & 45.5 & 69.8 & 43.6 & 20.7 & 63.5 & 85.5 & 62.7 & 41.1 & 62.7 & 84.5 & 61.9 & 40.3 \\
				\rowcolor{gray!20}
				& \Name(rv) & 60.1 & 84.1 & 61.8 & 35.7 & 59.7 & 83.6 & 61.3 & 35.2 & 74.2 & 89.8 & 74.7 & 54.8 & 73.5 & 89.2 & 74.0 & 53.9 \\
				\cline{2-18}
				& PointPillars* & 63.6 & 87.4 & 62.9 & 37.2 & 63.1 & 86.9 & 62.3 & 36.7 & 81.3 & 94.0 & 81.7 & 65.5 & 80.4 & 93.5 & 80.8 & 64.1 \\
				\rowcolor{gray!20}
				& \Name(pp) & 66.5 & 89.1 & 68.3 & 40.8 & 66.1 & 88.7 & 67.7 & 40.3 & 81.2 & 94.2 & 81.8 & 63.6 & 80.5 & 93.7 & 81.0 & 62.3 \\
				& \Name(2x) & 68.0 & 91.1 & 68.1 & 41.1 & 67.6 & 90.7 & 67.6 & 40.5 & 81.6 & 94.3 & 82.2 & 65.6 & 80.9 & 93.9 & 81.4  & 64.2 \\
				\rowcolor{gray!20}
				& \Name(2xc) & \textbf{68.3} & 91.3 & \textbf{68.5} & \textbf{42.4} & \textbf{67.9} & 90.9 & \textbf{ 68.0} & \textbf{41.8} & \textbf{81.7} & 94.3 & \textbf{82.3} & \textbf{65.8} & \textbf{81.0} & 93.9 & \textbf{81.5} & \textbf{64.5} \\
				\hline
			\end{tabular}
		}
	\end{center}
	\caption{Vehicle detection results on Waymo Open Dataset validation sequences. Here PointPillars~\cite{pointpillars} is reproduced by ~\cite{mvf} and PointPillar* is implemented in mmdetection3d. Our \Name(pp) results are based on PointPillars*' proposals. \Name(2x) means we double the channels in points encoding network and \Name(2xc) means we cascade another stage on the \Name(2x)'s outputs. The RV first stage is reproduced by ourselves following \cite{rcd}, but without the Range Conditioned Covolution operator because it is not open-sourced.}
	\label{tab:waymo-vehicle}
	\vspace{-4mm}
\end{table*}
\subsection{Size Ambiguity Problem} \label{subsec:Problem}

The point-based R-CNN is quite straight-forward. Similar models have already been exploited in previous works \cite{pointrcnn, fast-pointrcnn, ipod}. However, when we directly apply such an algorithm, a clear performance drop is observed. The drop is most significant when the class is of mixed size or multiple classes are trained together. After careful analysis, we find an intriguing size ambiguity problem in the point-based R-CNN models.

The problem is related to the property of point cloud. Unlike 2D images, in which each position is filled with RGB values, the point cloud is sparse and irregular. Nevertheless, if we directly use a point-based method, in which we only consider existing points in network, we would ignore the scale information indicated by the spacing. Taking Fig.~\ref{fig:solutions}(a) as an example, the dashes blue and green bounding boxes will have the same feature representation if we only consider the red points in it, however their classification and regression targets may differ a lot. Different from 2D RCNN, we should equip our LiDAR-RCNN with the ability to perceive the spacing and the size of proposals. In \secref{subsec:verify}, we show some statistics from real data. As a result, we propose the following size aware methods to enhance the features.

\subsection{Size Aware Point Features}
In this section, we elaborate several methods to fix the issue mentioned above. These methods will be discussed and compared in detail in the experiments. 

\noindent
\textbf{Normalization.} The simplest solution to the ambiguity problem is to normalize the point coordinates by the proposal size. Then the proposal's boundaries are aligned to $\{-\frac{1}{2}, \frac{1}{2}\}$. If the proposal is enlarged, the point coordinates will be smaller and the size target will be higher. Consequently, the model could be aware of the size of the proposal. However, size normalization makes the proposal to be a unit cube and the object shape will be distorted (Fig. \ref{fig:solutions}(b)). When the R-CNN model is applied on multiple categories, it totally ignores the scale difference off different categories. The size normalization makes it more difficult for the model to distinguish different categories.

\noindent
\textbf{Anchor.} One solution proposed in some previous works\cite{ipod, frustum} is to set an anchor for each category. Then the regression targets will be based on a fixed anchor, which eliminates the ambiguity of the size target. However, our goal is to judge the quality (whether the proposal's IoU w.r.t. the ground-truth is larger than a threshold) and refine the proposal, not the anchor. Since the network is still not aware of the boundary of the proposal, this method does not solve the classification ambiguity. Furthermore, when there are few point in the boxes, objects from different categories will also have similar features. In this case, there will be a regression ambiguity because different categories have different anchors, which corresponds to different regression targets.

\noindent
\textbf{Voxelization.} Several second stage detectors voxelize the points in proposal boxes\cite{parta2, rcd}. The empty voxels or point distribution in the voxels may indicate the size and spacing in the proposal as shown in Fig. \ref{fig:solutions}(d). However, in this solution, the voxel actually acts as a ``sub-anchor''. The points in it are still not aware of the voxel boundary. The model only have coarse information about the proposal size at voxel level, but not the point level. As a result, this solution alleviates, but not fully solves the ambiguity problem.

Revisiting the previous solutions, we can conclude that the key is to provide the size information to network, while preserving the shape of the object. To achieve this, we propose two candidate solutions that can solve the problem.


\noindent
\textbf{Boundary offset.} To provide the proposal boundary information, a simple way is to append the boundary offset to the point features. From the offset, the network will be able to know how far the points is from the proposal's boundary, which should solve the ambiguity problem.

\noindent
\textbf{Virtual points.} Since the R-CNN model ignores the spacing, another natural idea is to augment the spacing with ``virtual points'' to indicate the existence of them. Here we generate the grid points which are evenly distributed in the proposal. The number of virtual points for all proposals are equal and the coordinates of virtual points has the same normalization with the real points. We append one binary feature to all the points to indicate whether the points are real or virtual. Through the virtual points, the RCNN module will have the ability to perceive the size of the proposal, since the spacing is no longer under represented.

Actually there may be other alternative solutions. We believe they will also work as long as providing the proposal size information to the R-CNN model. The main contribution of this paper is to clarify the size ambiguity problem. The solution is not limited within the proposed ones.

\begin{table*}
	\begin{center}
		\scalebox{1}[1]{
			\setlength\tabcolsep{2pt}
			\begin{tabular}{c|c|cc|cc|cc}
				\hline
				& &
				\multicolumn{2}{c|}{vehicle (IoU=0.7)} & \multicolumn{2}{c|}{pedestrian (IoU=0.5)} &
				\multicolumn{2}{c}{cyclist (IoU=0.5)} \\
		    	\multirow{-2}{*}{Difficulty} & \multirow{-2}{*}{Method} & 3D AP  & 3D APH &
				3D AP & 3D APH  &
				3D AP & 3D APH \\
				\hline
				LEVEL\_1
				& SECOND~\cite{second} & 58.5 & 57.9 & 63.9 & 54.9 & 48.6 & 47.6  \\
				\rowcolor{gray!20}
				& \Name(sec) & 62.6 & 62.1 & 68.2 & 59.5 & 52.8 & 51.6 \\
				\cline{2-8}
				& PointPillars~\cite{pointpillars} & 71.6 & 71.0 & 70.6 & 56.7 & 64.4 & 62.3  \\
				\rowcolor{gray!20}
				& \Name(pp) & 73.4 & 72.9 & 70.6 & 57.8 & 66.8 & 64.8  \\
				
				& \Name(2x) & 73.5 & 73.0 & 71.2 & 58.7 & 68.6 & 66.9  \\
				\hline
				LEVEL\_2
				& SECOND~\cite{second} & 51.6 & 51.1 & 56.0 & 48.0 & 46.8 & 45.8  \\
				\rowcolor{gray!20}
				& \Name(sec) & 54.5 & 54.0 & 59.3 & 51.7 & 50.9 & 49.7 \\
				\cline{2-8}
				& PointPillars~\cite{pointpillars} & 63.1 & 62.5 & 62.9 & 50.2 & 61.9 & 59.9   \\
				\rowcolor{gray!20}
				& \Name(pp) & 64.6 & 64.1 & 62.5 & 50.9 & 64.3 & 62.4  \\
				
				& \Name(2x) & 64.7 & 64.2 & 63.1 & 51.7 & 66.1 & 64.4  \\
				\hline
			\end{tabular}
		}
	\end{center}

	\caption{Multi-class 3D detection results on Waymo Open Dataset validation sequences. Both PointPillars~\cite{pointpillars} and SECOND~\cite{second} baselines are implemented in mmdetection3d. 2x means we double the channels in points encoding network.}
	\label{tab:waymo-3d}
	\vspace{-3mm}

\end{table*}

\section{Experiments}
In this section, we first briefly introduce the datasets we use in \secref{subsec:datasets}, then in \secref{subsec:verify} we show some statistics to support our points in \secref{subsec:Problem}.The implementation details of our \Name is provided in \secref{subsec:Implementation Details}. \secref{subsec:Waymo Open Dataset} shows that \Name can generally improve the performance of all baseline models, and surpasses the previous state-of-the-art methods. In \secref{subsec:Ablation}, we conduct extensive ablation studies to show that the proposed methods can effectively solve the problem described in \secref{subsec:Problem}.

\subsection{Datasets}\label{subsec:datasets}
\textbf{Waymo Open Dataset(WOD)~\cite{wod}} is a recently released public dataset for autonomous driving research. It contains RGB images from five high-resolution cameras and 3D point clouds from five LiDAR sensors. 
The whole dataset consists of 1000 scenes(20s fragment) for training and validation and 150 scenes for testing. For the 3D detection task, the dataset has 12M labeled 3D LiDAR objects which are annotated in full 360-degree field, including 4.81M vehicles and 2.22M pedestrians for training and 1.25M vehicles and 539K pedestrians for validation. It is one of the largest and most diverse autonomous driving datasets.
For evaluation, Average precision (AP) and APH are used as the metric. Specifically, ~\cite{wod} proposes APH to incorporate heading information into detection metrics. The IoU threshold for vehicle detection is 0.7 while 0.5 for pedestrian and cyclist.

\textbf{KITTI dataset~\cite{kitti}} is one of the most widely used dataset in 3D detection task. It is equipped with sensors like GPS/IMU with RTK, stereo cameras, Velodyne HDL-64E. There are 7481 training samples which are commonly divided into 3712 samples for training and 3769 samples for validation, and 7518 samples for tesing. 

\subsection{Emperical Study of Scale Ambiguity Problem} \label{subsec:verify}
As explained in \secref{subsec:Problem}, the Size Ambiguity Problem means when the proposals are enlarged, the point clouds (also the features from PointNet) falling in them may not change. We verify our hypothesis on the WOD with a typical 3D detector, PointPillars~\cite{pointpillars}.
The total number of output bounding boxes from PointPillars in the WOD testing set is $4,387,747$. When we enlarge every box's width and length by $1$ meter, $21.3\%$ boxes keep the same number of the points, and $42.5\%$ boxes only get new points less than $10$, and most of them are ground points. 

The output features of these bounding boxes and their enlarged ones will be very similar, while their classification and regression targets significantly differ. These numbers reflect how ubiqutous the Scale Ambiguity Problem is.

\subsection{Implementation Details} \label{subsec:Implementation Details}
For all experiments except voxelization, we use the network architecture shown in \figref{fig:pointnet_backbone}. We shrink the number of embedding channels to $[64, 64, 512]$ in PointNet to achieve fast inference speed while maintaining accuracy. 512 points are sampled for each proposal. If the points number in one bounding box is fewer than 512, we randomly repeat the points to keep the input dimension.

For the voxelization experiments, we use two fully connected layer to extract features in each voxel. Then the features are max-pooled in voxels to form a $14\times 14 \times 14$ 3D voxel map. Finally, five 3D convolution layers and three fully-connected layers are utilized to extract the R-CNN features. 

To ensure the reproducibility of our algorithm, we use the code base of mmdetection3d\footnote{https://github.com/open-mmlab/mmdetection3d}
to extract proposals from Waymo Open Dataset. We directly save the outputs from PointPillars~\cite{pointpillars} and SECOND~\cite{second} with default testing settings in mmdetection3d as our inputs. There are about 4M vehicle proposals, 1.8M pedestrian proposals and 72K cyclist proposals for each model. Following PointRCNN~\cite{pointrcnn}, we randomly jitter the vehicle proposals during training as augmentation.

Besides the models from MMDetection3D, we also apply our proposed algorithm on the recently proposed RCD model~\cite{rcd}. It is a two-stage detector with a range view based first stage and a voxel-based second stage. Since it is not open-sourced, we only re-implemented its baseline model without the Range Conditioned Convolution operator and the Soft Range Gating. The results of both stages match with the original paper.

In our experiments, all models are trained using SGD with momentum of 0.9 for 60 epochs. The training schedule is a poly learning rate update strategy, starting from 0.02 with $10^{-5}$ weight decay. Training the model on WOD takes 7 hours on 4 Nvidia Geforce GTX2080Ti GPU with 256 batch size per GPU. 


\subsection{3D Detection on Open Datasets} \label{subsec:Waymo Open Dataset}
\textbf{Waymo Open Dataset.} We apply the proposed \Name to three different kinds of baseline models: PointPillars~\cite{pointpillars}, SECOND~\cite{second}, and RV~\cite{rcd} which are birds-eye-view based, 3D voxel based, and range view based, respectively. All these methods are evaluated with 3D AP in two difficulty levels defined in the official evaluation, where the LEVEL\_1 objects should contain at least 5 points in each groud-truth bounding box while the LEVEL\_2 objects only need one single point.

~\tabref{tab:waymo-vehicle} reports the results of vehicle detection with 3D and BEV AP on validation sequences. Note that after hyperparameters finetuning, the performance of PointPillars~\cite{pointpillars} implemented by MMDetection3D already surpasses the previous state-of-the-art model PV-RCNN~\cite{pv-rcnn}. As shown in ~\tabref{tab:waymo-vehicle}, our method improves all the baseline models with a significant margin. Benchmark on various distance ranges shows that \Name generally improves the baseline consistently on all distances. Furthermore, with the strong PointPillars baseline, \Name outperforms the PV-RCNN by $5.31\%$ AP on the 3D vehicle detection task. We also experiment with a 2x model with double channels, which enhances performance on the more challenging case, $1.5\%$ improvement in the LEVEL\_2. Based on this, we cascade another stage to show the ability of our models, which increase the 3D AP to $76\%$ on vehicle detetion. For the range view model, without bells and whistles, we outperform the RCD~\cite{rcd}.

Additionally, we show multi-class detection results with VEHICLE, PEDESTRIAN, and CYCLIST in ~\tabref{tab:waymo-3d}. Because we only obtain the multi-class results of PointPillars~\cite{pointpillars} and SECOND~\cite{second} from publicly available codes, we test our method with these two baselines only. We use a multi-class classification loss and a class-agnostic regression loss for the three classes detection, without any other tuning. ~\tabref{tab:waymo-3d} demonstrates that our model also performs well in detecting small objects besides vehicles.

\begin{table}[t]
	\centering
	\setlength{\tabcolsep}{6pt}
	\begin{tabular}{l|c|c|c}
		\hline
		Methods    & Easy & Moderate  & Hard \\ \midrule
		PointPillars(T)  & 82.17 & 72.81 & 67.57 \\
		PointPillars(TV)  & 83.12 & 74.11 & 69.14 \\
		Ours(TV)      & 85.97 &  74.21 & 69.18 \\
		\hline
	\end{tabular}
	\vspace{2mm}
	\caption{Vehicle 3D AP results on KITTI testing set. T means trained with training set. TV means trained with training+validation set.}
	\label{tab:KITTI}
\end{table}

\textbf{KITTI Dataset.} We also show the effectiveness of the proposed method on KITTI dataset~\cite{kitti}. LiDAR R-CNN works by fitting the residuals upon first-stage models. However, models can easily achieve $>98\%$ AP (easy set) on the small training set of KITTI with only $3,712$ frames. As a result, the limited training data reduces the effectiveness of our LiDAR R-CNN. As shown in Tab.~\ref{tab:KITTI}, we use PointPillars(T) as the first-stage and train LiDAR R-CNN with trainval data to get more effective data. We also show PointPillars(TV) to make a fair comparison. Though the improvement is less than those on WOD, we still highlight 2.8 AP improvement on easy set.

\subsection{Ablation Studies} \label{subsec:Ablation}
In this section, we first conduct an ablation study to support the claims on \secref{subsec:Problem}. And then, we show the generalization and inference time of proposed method. We show both single-class and multi-class results in Waymo Open Dataset in this section.

\textbf{The size ambiguity problem}.
\begin{table}[]
	\centering
	\setlength{\tabcolsep}{6pt}

	\begin{tabular}{l|c|c}
		\toprule
		Methods    & 3D AP@70 & BEV AP@70  \\ \midrule
		PointPillars~\cite{pointpillars} & 71.6 & 87.1  \\
		PointNet refinement & 74.1 & 87.9  \\
		voxel    & 72.9 & 87.2   \\
		anchor    & 75.2 & 88.2   \\
		size normalization    & 75.4 &  88.1   \\
		\midrule
		virtual point   & 75.4 &  88.1   \\
		boundary offset      & 75.6 &  88.3   \\
		\bottomrule
	\end{tabular}
	\vspace{2mm}
	\caption{Ablation studies for different scale-aware methods proposed in \secref{subsec:Problem} on Waymo Open Dataset vehicle detection. The baseline uses a basic PointNet with [x,y,z] inputs as second-stage on PointPillars. }
	\label{tab:ablation-single}
\end{table}
\tabref{tab:ablation-single} shows the comparison with different methods proposed in \secref{subsec:Problem}. From the table, we can conclude that all the point-based R-CNN models outperforms the baseline PointPillars. The PointNet models performs even better than voxel based models. All other solutions achieve similar improvements compared with baseline on the one-class setting.

\begin{table}[]
	\centering
	\setlength{\tabcolsep}{6pt}
	\begin{tabular}{l|c|c|c}
		\toprule
		Methods    & vehicle & pedestrian  & cyclist \\ \midrule
		PointPillars~\cite{pointpillars}  & 71.6 & 70.6 & 64.4 \\
		PointNet refinement   & 72.1 & 69.2 & 62.2 \\
		voxel   & 72.1 & 69.8 & 64.5 \\
		anchor   & 72.5 & 70.2 & 63.5 \\
		size normalization    & 72.7 & 69.9 & 64.4   \\
		\midrule
		virtual point   & 73.3 &  70.4 & 66.2  \\
		boundary offset      & 73.4 &  70.6 & 66.8  \\
		\bottomrule
	\end{tabular}
	\vspace{2mm}
	\caption{3D AP results on WOD with three classes trained in one model.}
	\label{tab:ablation-three}
\end{table}

Nevertheless, when it comes to the multi-class problem, the size normalization and anchor solution suffer from difficulty in distinguishing different categories as shown in \tabref{tab:ablation-three}. Normalization using proposals' size will make proposals from different classes have similar scale, which would confuse the R-CNN model. When the network is not sure about the class of the input objects, it will also be confused about which anchor should be based on. The virtual point and offset methods, which fundamentally solve the ambiguity problem, reach the best results on the multi-class setting. 

\begin{table}[]
	\centering
	\setlength{\tabcolsep}{6pt}
	\begin{tabular}{l|c|c}
		\toprule
		Methods    & 3D AP@70 & BEV AP@70  \\ \midrule
		RV baseline & 53.4 & 70.5 \\
		RV + voxel & 64.1 & 76.7  \\
		RV+ \Name & \textbf{68.7} &  \textbf{81.1}   \\
		\bottomrule
	\end{tabular}
	\vspace{2mm}
	\caption{Comparison between voxel-based second-stage and proposed \Name on a range-view-based detector (RV). The RV baseline is re-implemented from the baseline of RCD~\cite{rcd}. The RV+\Name replaces the second-stage with our method.  }
	\vspace{-2mm}
	\label{tab:ablation-voxel}
\end{table}

In \tabref{tab:ablation-voxel}, we show the performance improvement of our \Name based on a range-view-based model RV. The RV model is the baseline model in RCD~\cite{rcd}. It is built with a voxel-based second stage, whose features are not only from the raw points, but also from the high level CNN feature maps. Similar with the voxel experiments in \tabref{tab:ablation-single}, the voxelization method is inferior to our purely point-based R-CNN model. 




\begin{figure}[t]
	\centering
	\includegraphics[width=0.9\linewidth]{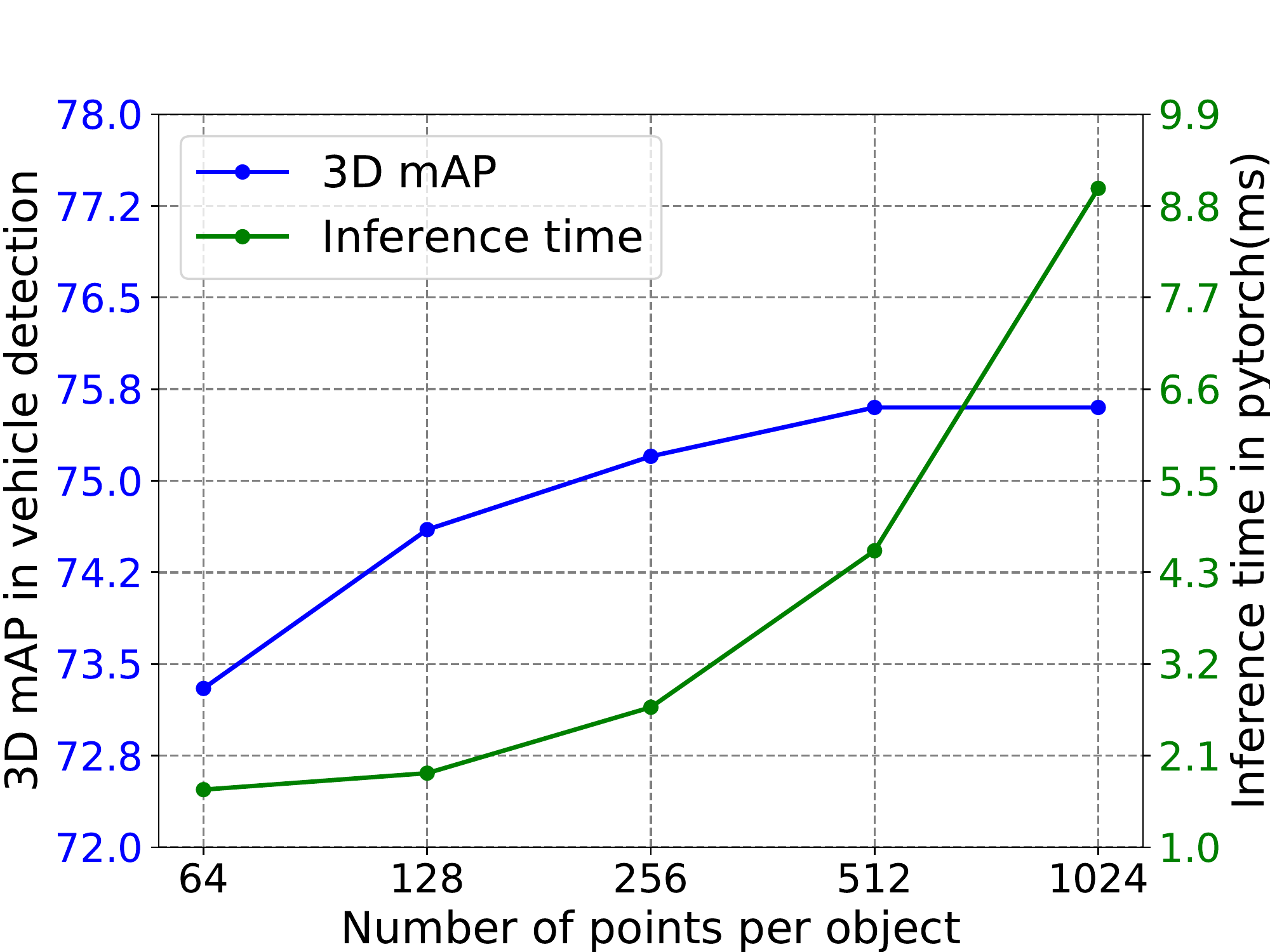}
	\caption{Adaptive inference time with different sampled points. Evaluate the \Name sampling 64 to 1024 points per object on Waymo Open Dataset Validation set.}
	\label{fig:number_points}
\end{figure}

\textbf{Network volume}.
A trade-off between speed and accuracy can be achieved by modifying the channels in points encoding. To quantify the effect of channel numbers, we show a \Name(2x) in \tabref{tab:waymo-3d} which obtains further $0.6\%$ improvement in pedestrian and $1.8\%$ in cyclist with two times channels. In addition, we also show the adaptive inference time by changing the number of sampled points per objects in \figref{fig:number_points}.

\textbf{Cross-model generalization}
\begin{table}[]
	\centering
	\setlength{\tabcolsep}{6pt}
	\begin{tabular}{l|c|c|c}
		\toprule
		Methods    & vehicle & pedestrian  & cyclist \\ \midrule
		SECOND~\cite{second}   & 58.5 & 63.9 & 48.6 \\
		\Name(sec)    & 62.6 & 68.2 & 52.8   \\
		\midrule
		\Name(pp)   & 62.1 &  67.2 & 52.9  \\
		\bottomrule
	\end{tabular}
	\vspace{2mm}
	\caption{We show generalization of our \Name in this Table. \Name(sec) use the outputs from SECOND for training and testing. \Name(pp) trains with PointPillars' outputs while inferencing on SECOND without finetuning.}
	\vspace{-2mm}
	\label{tab:ablation-generalization}
\end{table}
\tabref{tab:ablation-generalization} demonstrates that our approach is able to generalize across models. Without finetuning, \Name trained on PointPillars~\cite{pointpillars} works well on SECOND~\cite{second} model.

\textbf{Sub-task ablation}
\begin{table}
	\small 
	\begin{center}
		\scalebox{1}[1]{
			\setlength\tabcolsep{2pt}
			\begin{tabular}{l|cccc|cc}
				\hline
				Method & 
				\tabincell{c}{score} & 
				\tabincell{c}{center} &
				\tabincell{c}{size} &
				\tabincell{c}{heading} 
				& AP@L1 & AP@L2 \\
				\hline
				PointPillars~\cite{pointpillars} &  & & & &  72.1 & 63.6  \\ \midrule
				PointNet & \checkmark & & &   & 72.9 & 64.2   \\
				refinement&  & \checkmark & &  & 72.9 & 64.3  \\
				&  &  & \checkmark&  &  72.2 & 63.7  \\
				&  &  & & \checkmark &  72.1 & 63.6  \\ 
				&\checkmark  & \checkmark & \checkmark& \checkmark &  74.1 & 65.3  \\ \midrule
				\Name & \checkmark & & &   & 73.3 & 64.5   \\
				boudary offset&  & \checkmark & &  &  73.7 & 65.0  \\
				&  &  & \checkmark&  &  72.6 & 64.1  \\
				&  &  & & \checkmark &  72.2 & 63.6  \\
				&\checkmark  & \checkmark & \checkmark& \checkmark &  75.6 & 66.5  \\ 
				\hline
			\end{tabular}
		}
	\end{center}
	\caption{Sub-task ablation study.}
	\label{tab:ablation-real}
\end{table}
An R-CNN model contains two tasks, namely classification task and regression task. The regression task can be further splitted into three sub-tasks, center regression, size regression and orientation regression. In \tabref{tab:ablation-real}, we successively add these sub-task refinement on the baseline model to see the effect of each sub-task. From the table, we can infer that the vanilla PointNet refinement cannot improve the performance of the size sub-task based on the PointPillars\cite{pointpillars} model. With the boundary offset, our \Name has no ambiguity problem and is able to refine the size. 

\begin{table}[]
	\centering
	\setlength{\tabcolsep}{6pt}

	\begin{tabular}{l|c|c|c}
		\toprule
		Methods    & 3D AP & time  & Params \\ \midrule
		voxel   & 72.9 & 19ms & 3.5M \\
		ours  & 75.6 & 4.5ms & 0.5M   \\
		\bottomrule
	\end{tabular}
	\vspace{2mm}
	\caption{Inference time comparison between voxel-based and our proposed method. The inference time is evaluated with 128 proposals(batchsize=128).}
	\vspace{-2mm}
	\label{tab:ablation-speed}
\end{table}


Finally, we show our inference time and model size in \tabref{tab:ablation-speed}. Memory and computational demands of \Name are marginal.

\section{Conclusions and Future Work}
This paper presents \NameNS, a fast and accurate second-stage 3D detector. Through a detailed analysis of scale ambiguity problem and deliberate experiments, we come up with practical solutions. Comprehensive experiments on Waymo Open Dataset demonstrate our method can improve steadily on all baseline models and state-of-the-art performance.

A multi-sensor fusion perception system is necessary for robotics and autonomous driving. Besides the encouraging performance on single-frame LiDAR detection, our \Name is easy to generalize to other kinds of inputs such as multi-frame LiDAR and RGB+LiDAR. As a second-stage framework, our method is more comfortable with various aggregated inputs. We will investigate how to develop our method into a multimodal fusion framework.

{\small
\bibliographystyle{ieee_fullname}
\bibliography{egbib}
}

\end{document}